\documentclass{article}
\usepackage[final, nonatbib]{nips_2016}

\usepackage[T1]{fontenc}
\usepackage{tgtermes}
\usepackage{amsmath}
\usepackage{enumitem}

\usepackage{amssymb}
\newcommand{\mathbold}[1]{\ensuremath{\boldsymbol{\mathbf{#1}}}}

\usepackage{scalefnt,letltxmacro}
\LetLtxMacro{\oldtextsc}{\textsc}
\renewcommand{\textsc}[1]{\oldtextsc{\scalefont{1.10}#1}}
\usepackage[ttdefault=true]{AnonymousPro}

\usepackage[usenames,dvipsnames]{xcolor}
\definecolor{shadecolor}{gray}{0.9}

\usepackage[english]{babel}
\usepackage[parfill]{parskip}
\usepackage{afterpage}
\usepackage{framed}

{\endMakeFramed}

\DeclareRobustCommand{\parhead}[1]{\textbf{#1}~}

\usepackage{lineno}

\usepackage{ragged2e}

\newcounter{parcount}

\usepackage{graphicx}
\usepackage[labelfont=bf]{caption}
\usepackage[format=hang]{subcaption}

\usepackage{booktabs}

\usepackage[algoruled]{algorithm2e}
\usepackage{listings}
\usepackage{fancyvrb}
\fvset{fontsize=\normalsize}

\usepackage[numbers]{natbib}

\usepackage[colorlinks,linktoc=all]{hyperref}
\hypersetup{citecolor=Blue}
\hypersetup{linkcolor=MidnightBlue}
\hypersetup{urlcolor=MidnightBlue}

\newcommand{\myeqp}[1]{\hyperref[eq:#1]{Eq.\ref*{eq:#1}}}
\newcommand{\mysub}[1]{\hyperref[sub:#1]{Section~\ref*{sub:#1}}}
\newcommand{\mysec}[1]{\hyperref[sec:#1]{Section~\ref*{sec:#1}}}
\newcommand{\mytable}[1]{\hyperref[table:#1]{Table~\ref*{table:#1}}}
\newcommand{\myfig}[1]{\hyperref[fig:#1]{Figure~\ref*{fig:#1}}}
\newcommand{\myappendix}[1]{\hyperref[appendix:#1]{Appendix~\ref*{appendix:#1}}}
\newcommand{\myalg}[1]{\hyperref[alg:#1]{Algorithm~\ref*{alg:#1}}}
\newcommand{\mytheorem}[1]{\hyperref[theorem:#1]{Theorem~\ref*{theorem:#1}}}
\newcommand{\myfootnote}[1]{\hyperref[footnote:#1]{Footnote~\ref*{footnote:#1}}}

\usepackage
[acronym,smallcaps,nowarn,section,nonumberlist]{glossaries}
\glsdisablehyper{}

\lstdefinestyle{mystyle}{
    commentstyle=\color{OliveGreen},
    keywordstyle=\color{BurntOrange},
    numberstyle=\tiny\color{black!60},
    stringstyle=\color{MidnightBlue},
    basicstyle=\ttfamily,
    breakatwhitespace=false,
    breaklines=true,
    captionpos=b,
    keepspaces=true,
    numbers=left,
    numbersep=5pt,
    showspaces=false,
    showstringspaces=false,
    showtabs=false,
    tabsize=2
}
\usepackage{tikz}

\usetikzlibrary{bayesnet}

\pgfdeclarelayer{edgelayer}
\pgfdeclarelayer{nodelayer}
\pgfsetlayers{edgelayer,nodelayer,main}

\definecolor{hexcolor0xbfbfbf}{rgb}{0.749,0.749,0.749}

\tikzset{>=latex}
\tikzstyle{none}   = [inner sep=0pt]
\tikzstyle{line}   = [ -, thick, shorten <=1pt, shorten >=1pt ]
\tikzstyle{arrow}  = [ ->, thick, shorten <=1pt, shorten >=1pt ]
\tikzstyle{ardash} = [ dashed, ->, thick, shorten <=1pt, shorten >=1pt ]

\tikzstyle{empty}=[circle,opacity=0.0,text opacity=1.0,inner sep=0pt]
\tikzstyle{box}=[rectangle,fill=White,draw=Black]
\tikzstyle{filled}=[circle,thick,fill=hexcolor0xbfbfbf,draw=Black]
\tikzstyle{hollow}=[circle,thick,fill=White,draw=Black]
\tikzstyle{param}=[rectangle,fill=Black,draw=Black,inner sep=0pt,minimum width=4pt,minimum height=4pt]
\tikzstyle{paramhollow}=[rectangle,thick,fill=White,draw=Black,inner sep=0pt,minimum
width=4pt,minimum height=4pt]

\usepackage{pgfplots}                               
\pgfplotsset{compat=newest}
\pgfplotsset{plot coordinates/math parser=false}
\newlength\figureheight
\newlength\figurewidth
\newlength\figureheightsmall
\newlength\figurewidthsmall
\definecolor{POSTcolor}{rgb}{0.48, 0.20, 0.58} 
\definecolor{Qcolor}{rgb}{0.00, 0.53, 0.22} 

\DeclareRobustCommand{\mb}[1]{\mathbold{#1}}

\newcommand{\mbb}{\mathbold{b}}

\newcommand{\mbh}{\mathbold{h}}

\newcommand{\mbw}{\mathbold{w}}
\newcommand{\mbx}{\mathbold{x}}
\newcommand{\mby}{\mathbold{y}}
\newcommand{\mbz}{\mathbold{z}}

\newcommand{\mbI}{\mathbold{I}}

\newcommand{\mbW}{\mathbold{W}}

\newcommand{\mbepsilon}{\mathbold{\epsilon}}

\newcommand{\mblambda}{\mathbold{\lambda}}

\newcommand{\mbtheta}{\mathbold{\theta}}

\newcommand{\E}{\mathbb{E}}

\newcommand{\cL}{\mathcal{L}}
\newcommand{\cF}{\mathcal{F}}

\newcommand{\cQ}{\mathcal{Q}}

\newcommand{\g}{\, | \,}
\renewcommand{\gg}{\,\|\,}

\usepackage{amsthm}
\newtheorem{theorem}{Theorem}

\newacronym{KL}{kl}{Kullback-Leibler}
\newacronym{ELBO}{elbo}{evidence lower bound}
\newacronym{EP}{ep}{expectation propagation}

\newacronym{MC}{mc}{Monte Carlo}
\newacronym{MCMC}{mcmc}{Markov chain Monte Carlo}

\newacronym{VI}{vi}{variational inference}
\newacronym{MFVI}{mfvi}{mean-field variational inference}
\newacronym{SVI}{svi}{stochastic variational inference}

\newacronym{VMP}{vmp}{variational message passing}

\newacronym{ADVI}{advi}{automatic differentiation variational inference}

\newacronym{RMSPROP}{rmsprop}{rmsprop}

\newacronym{NUTS}{nuts}{no-U-turn sampler}
\newacronym{HMC}{hmc}{Hamiltonian Monte Carlo}

\newacronym{ARD}{ard}{automatic relevance determination}
\newacronym{GMM}{gmm}{Gaussian mixture model}
\newacronym{DLGM}{DLGM}{deep latent Gaussian model}
\newacronym{LS}{ls}{Langevin-Stein}
\newacronym{OPVI}{opvi}{operator variational inference}

\title{Operator Variational Inference}

\author{%
Rajesh Ranganath \\
Princeton University \\
\And
Jaan Altosaar \\
Princeton University\\
\And
Dustin Tran \\
Columbia University \\
\And
David M.~Blei \\
Columbia University
}

\begin{document}
\maketitle

\begin{abstract}
Variational inference is an umbrella term for algorithms which cast
Bayesian inference as optimization.
Classically, variational inference uses the Kullback-Leibler
divergence to define the optimization.
Though
this divergence has been widely used, the resultant posterior approximation
can suffer from undesirable statistical properties. To address this, we
reexamine
variational inference from its roots as an optimization problem. We use
\textit{operators}, or functions of functions, to design variational
objectives.
As one example,
we design a
variational objective
with a Langevin-Stein operator.
We develop
a black box algorithm, \gls{OPVI},
for optimizing any operator objective.
Importantly, operators enable us to make explicit the statistical and
computational tradeoffs for variational inference.
We can characterize different properties of variational objectives, such as
objectives that admit \emph{data
subsampling}---allowing inference to scale to massive data---as well as
objectives that admit
\emph{variational programs}---a rich class of posterior approximations that
does not require a tractable density.
We illustrate the benefits of \gls{OPVI} on a
mixture model and a generative model of images.
\end{abstract}

\section{Introduction}
\label{sec:introduction}

Variational inference is an umbrella term for algorithms that cast
Bayesian inference as optimization~\citep{Jordan:1999}.  Originally
developed in the 1990s, recent advances in variational inference have
scaled Bayesian computation to massive
data~\citep{Hoffman:2013}, provided black box strategies
for generic inference in many models~\citep{Ranganath:2014}, and
enabled more accurate approximations of a model's posterior without
sacrificing efficiency~\citep{Rezende:2015,
  ranganath2016hierarchical}.  These innovations have both scaled
Bayesian analysis and removed the analytic burdens that have
traditionally taxed its practice.

Given a model of latent and observed variables $p(\mbx, \mbz)$, variational
inference posits a family of distributions over its latent variables
and then finds the member of that family closest to the posterior,
$p(\mbz \g \mbx)$. This is typically formalized as minimizing a \gls{KL}
divergence from the approximating family $q(\cdot)$ to the posterior
$p(\cdot)$.  However, while the $\textsc{kl}(q \gg p)$ objective offers
many beneficial computational properties, it is ultimately designed
for convenience; it sacrifices many desirable statistical properties
of the resultant approximation.

When optimizing \gls{KL}, there are two issues with the posterior
approximation that we highlight.  First, it typically underestimates the
variance of the posterior.  Second, it can result in degenerate
solutions that zero out the probability of certain configurations of
the latent variables.  While both of these issues can be partially
circumvented by using more expressive approximating families, they
ultimately stem from the choice of the objective. Under the \gls{KL}
divergence, we pay a large price when $q(\cdot)$ is
big where $p(\cdot)$ is tiny; this price becomes infinite when $q(\cdot)$ has
larger support than $p(\cdot)$.

In this paper, we revisit variational inference from its core
principle as an optimization problem. We use
\textit{operators}---mappings from functions to functions---to design
variational objectives, explicitly trading off computational
properties of the optimization with statistical properties of the
approximation.  We
use operators to formalize the basic properties
needed for variational inference algorithms. We
further outline how to use them to define new
variational objectives; as one example, we
design a variational objective using a Langevin-Stein operator.

We develop \glsreset{OPVI}\gls{OPVI}, a black box algorithm that optimizes any
operator objective.  In the context of \gls{OPVI}, we show that the
Langevin-Stein objective enjoys two good properties.  First, it is
amenable to \textit{data subsampling}, which allows inference to scale to
massive data.  Second, it permits rich approximating families, called
\emph{variational programs}, which do not require analytically tractable
densities. This greatly expands the class of variational families and
the fidelity of the resulting approximation. (We note that the
traditional \gls{KL} is not amenable to using variational programs.)
We study \gls{OPVI} with the Langevin-Stein objective on a
mixture model and a generative model of images.

\parhead{Related Work.}  There are several threads of research in
variational inference with alternative divergences.  An early example
is \gls{EP}~\citep{minka2001expectation}.  \gls{EP} promises
approximate minimization of the inclusive \gls{KL} divergence
$\textsc{kl}(p || q)$ to find overdispersed approximations to the
posterior.  \gls{EP} hinges on local minimization with respect to
subsets of data and connects to work on $\alpha$-divergence
minimization~\citep{minka2004power, hernandezlobato2015black}.
However, it does not have convergence guarantees and typically does not
minimize \gls{KL} or an $\alpha$-divergence because it is
not a global optimization method. We note that these divergences can be
written as operator variational objectives, but they do not satisfy
the tractability criteria and thus require further approximations.
\citet{li2016r} present a variant of $\alpha$-divergences that satisfy
the full requirements of \gls{OPVI}.
Score matching~\citep{hyvarinen2005estimation}, a method for
estimating models
by matching the score function of one distribution to another that can
be sampled, also falls into the class of objectives we develop.

Here we show how to construct new objectives, including
some not yet studied. We make explicit the requirements to construct
objectives for variational inference. Finally, we discuss further
properties that make them amenable to both scalable and flexible
variational inference.

\section{Operator Variational Objectives}

We define operator variational objectives and the conditions needed
for an objective to be useful for variational inference. We develop a
new objective, the Langevin-Stein objective, and show how to place
the classical \gls{KL} into this class.  In the next section, we
develop a general algorithm for optimizing operator variational
objectives.

\subsection{Variational Objectives}

Consider a probabilistic model $p(\mbx, \mbz)$ of data $\mbx$ and
latent variables $\mbz$.  Given a data set $\mbx$, approximate Bayesian
inference seeks to approximate the posterior distribution
$p(\mbz \g \mbx)$, which is applied in all downstream tasks.
Variational inference posits a family of approximating distributions
$q(\mbz)$ and optimizes a divergence function to find the member of
the family closest to the posterior.

The divergence function is the \textit{variational objective}, a
function of both the posterior and the approximating
distribution. Useful variational objectives hinge on two properties:
first, optimizing the function yields a good posterior approximation;
second, the problem is tractable when the posterior distribution is
known up to a constant.

The classic construction that satisfies these properties is the \gls{ELBO},
\begin{align}
  \mathbb{E}_{q(\mbz)}[\log p(\mbx, \mbz) - \log q(\mbz)].
  \label{eq:elbo}
\end{align}
It is maximized when $q(\mbz)=p(\mbz \g \mbx)$ and it only depends on
the posterior distribution up to a tractable constant,
$\log p(\mbx, \mbz)$.  The \gls{ELBO} has been the focus in much of
the classical literature.  Maximizing the \gls{ELBO} is equivalent to
minimizing the \gls{KL} divergence to the posterior, and the
expectations are analytic for a large class of
models~\citep{Ghahramani:2001}.

\subsection{Operator Variational Objectives}

We define a new class of variational objectives, \textit{operator
  variational objectives}.  An operator objective has three
components.  The first component is an operator $O^{p,q}$ that depends
on $p(\mbz\g \mbx)$ and $q(\mbz)$. (Recall that an operator maps functions to
other functions.)  The second component is a family of test functions
$\cF$, where each $f(z) \in \cF$ maps realizations of the latent
variables to real vectors $\mathbb{R}^d$.  In the objective, the
operator and a function will combine in an expectation
$\E_{q(\mbz)}[(O^{p,q}\, f )(\mbz)]$, designed such that values close to zero
indicate that $q$ is close to $p$.  The third component is a distance function
$t(a):\mathbb{R} \rightarrow [0, \infty)$, which is applied to the
expectation so that the objective is non-negative. (Our example
uses the square function $t(a)=a^2$.)

These three components combine to form the operator variational
objective.  It is a non-negative function of the variational
distribution,
\begin{align}
  \cL(q ; O^{p,q}, \cF, t) =
  \sup_{f \in \cF}
  t(\E_{q(\mbz)}[(O^{p,q} \, f)(\mbz)]).
  \label{eq:operator-obj}
\end{align}
Intuitively, it is the worst-case expected value among all
test functions $f\in\cF$.  Operator variational inference seeks to minimize
this objective with respect to the variational family $q\in\cQ$.

We use operator objectives for posterior inference.  This
requires two conditions on the operator and function family.
\begin{enumerate}[leftmargin=*]
\item \emph{Closeness}.  The minimum of the variational objective is
  at the posterior, $q(\mbz)=p(\mbz\g \mbx)$.  We meet this condition by
  requiring that $\E_{p(\mbz\g \mbx)}[(O^{p,p} \, f)(\mbz)]=0$ for all $f \in \cF$.
  Thus, optimizing the objective will produce $p(\mbz\g \mbx)$ if it is the
  only member of $\cQ$ with zero expectation (otherwise it will
  produce a distribution in the equivalence class: $q \in \cQ$ with
  zero expectation).  In practice, the minimum will be the closest
  member of $\cQ$ to $p(\mbz \g \mbx)$.

\item \emph{Tractability}.  We can calculate the variational objective
  up to a constant without involving the exact posterior
  $p(\mbz\g \mbx)$.  In other words, we do not require calculating the
  normalizing constant of the posterior, which is typically
  intractable. We meet this condition by requiring that the operator
  $O^{p,q}$---originally in terms of $p(\mbz\g \mbx)$ and
  $q(\mbz)$---can be written in terms of $p(\mbx, \mbz)$ and
  $q(\mbz)$.  Tractability also imposes conditions on $\cF$: it must
  be feasible to find the supremum.  Below, we satisfy this by
  defining a parametric family for $\cF$ that is amenable to
  stochastic optimization.

\end{enumerate}
\myeqp{operator-obj} and the two conditions provide a mechanism to
design meaningful variational objectives for posterior inference.
Operator variational objectives try to match expectations with respect
to $q(\mbz)$ to those with respect to $p(\mbz\g \mbx)$.

\subsection{Understanding Operator Variational Objectives}

Consider operators where $\E_{q(\mbz)}[(O^{p,q} \, f)(\mbz)]$ only
takes positive values. In this case, distance to zero can be measured
with the identity $t(a)=a$, so tractability implies the operator need
only be known up to a constant.  This family includes tractable forms
of familiar divergences like the \gls{KL} divergence (\gls{ELBO}), R\'enyi's
$\alpha$-divergence~\citep{li2016r}, and the
$\chi$-divergence~\citep{nielsen2013chi}.

When the expectation can take positive or negative values, operator
variational objectives are closely related to Stein
divergences~\citep{Barbour:1988}. Consider a family of scalar test
functions $\cF^*$ that have expectation zero with respect to the
posterior, $\E_{p(\mbz \g \mbx)}[f^*(\mbz)] = 0$.  Using this family,
a \textit{Stein divergence} is
\begin{align*}
  D_{\textrm{Stein}}(p, q) =
  \sup_{f^* \in \cF^*} |\E_{q(\mbz)}[f^*(\mbz)] - \E_{p(\mbz\g \mbx)}[f^*(\mbz)]|.
\end{align*}
Now recall the operator objective of \myeqp{operator-obj}.  The
closeness condition implies that
\begin{align*}
  \cL(q ; O^{p,q}, \cF, t) = \sup_{f \in \cF}
  t(\E_{q(\mbz)}[(O^{p,q} \, f)(\mbz)] - \E_{p(\mbz \g \mbx)}[(O^{p,p} \, f)(\mbz)]).
\end{align*}
In other words, operators with positive or negative expectations lead
to Stein divergences with a more generalized notion of distance.

\subsection{Langevin-Stein Operator Variational Objective}

We developed the operator variational objective.  It is a class of
tractable objectives, each of which can be optimized to yield an
approximation to the posterior.  An operator variational objective is
built from an operator, function class, and distance function to zero.
We now use this construction to design a new type of variational
objective.

An operator objective involves a class of functions that has known
expectations with respect to an intractable distribution.  There are
many ways to construct such classes~\citep{Assaraf:1999,
  Barbour:1988}.  Here, we construct an operator objective from the
generator Stein's method applied to the Langevin diffusion.

Let $\nabla^\top f$ denote the divergence of a vector-valued
function $f$, that is, the sum of its individual gradients.
Applying the generator method of~\citet{Barbour:1988} to Langevin diffusion
gives the operator
\begin{align}
  (O^{p}_\textsc{ls} \, f)(\mbz) =  \nabla_z \log p(\mbx, \mbz)^\top f(\mbz) + \nabla^\top f.
  \label{eq:lang-stein}
\end{align}
We call this the \gls{LS} operator. See also \citet{mira2013zero, Oates:2014, Gorham:2015} for related derivations. We obtain the corresponding
variational objective by using the squared distance function and
substituting \myeqp{lang-stein} into~\myeqp{operator-obj},
\begin{align}
  \label{eq:lang-stein-objective}
  \cL(q ; O^{p}_\textsc{ls}, \cF) = \sup_{f \in \mathcal{F}}
  (\E_q[\nabla_z \log p(\mbx,
  \mbz)^\top f(\mbz) + \nabla^\top f])^2.
\end{align}
The \gls{LS} operator satisfies both conditions.  First, it satisfies
closeness because it has expectation zero under the posterior
(Appendix A) and its unique minimizer is the posterior (Appendix B).
Second, it is tractable because it requires only the joint
distribution. The functions $f$ will also be a parametric family, which we
detail later.

Additionally, while the \gls{KL} divergence finds variational
distributions that underestimate the variance, the \gls{LS} objective
does not suffer from that pathology.  The reason is that \gls{KL} is
infinite when the support of $q$ is larger than $p$; here this is not
the case.

We provided one example of a variational
objectives using operators, which is specific to continuous variables. In
general, operator objectives are not
limited to continuous variables; Appendix C describes an operator for
discrete variables.

\subsection{The KL Divergence as an Operator Variational Objective}
Finally, we demonstrate how classical variational methods fall inside
the operator family.  For example, traditional variational inference
minimizes the \gls{KL} divergence from an approximating family to the
posterior~\citep{Jordan:1999}. This can be construed as an operator
variational objective,
\begin{align}
  \label{eq:KL-operator}
  (O^{p,q}_{\rm KL}  \,  f)(\mbz) = \log q(\mbz) - \log p(\mbz | \mbx)
  \quad
  \forall f\in\cF.
\end{align}
This operator does not use the family of functions---it trivially maps
all functions $f$ to the same function. Further, because \gls{KL} is
strictly positive, we use the identity distance $t(a) = a$.

The operator satisfies both conditions.  It satisfies closeness
because ${\rm KL}(p || p) = 0$.  It satisfies tractability because it
can be computed up to a constant when used in the operator objective
of \myeqp{operator-obj}.  Tractability comes from the fact that
$\log p(\mbz \g \mbx) = \log p(\mbz, \mbx) - \log p(\mbx)$.

\section{Operator Variational Inference}
\label{sec:inference}
\glsresetall

We described operator variational objectives, a broad class of
objectives for variational inference.
We now examine how it can be optimized.
We develop a black box algorithm~\citep{Wingate:2013, Ranganath:2014}
based on Monte Carlo estimation and stochastic optimization.
Our algorithm applies to a general class of models and any
operator objective.

Minimizing the operator objective involves two optimizations:
minimizing the objective with respect to the approximating family
$\cQ$ and maximizing the objective with respect to the function class
$\cF$ (which is part of the objective).

We index the family $\cQ$
with \textit{variational parameters} $\lambda$ and require that it
satisfies properties typically assumed by black box
methods~\citep{Ranganath:2014}: the variational distribution $q(\mbz ;
\lambda)$ has a known and tractable density; we can sample from
$q(\mbz ; \lambda)$; and we can tractably compute the score function
$\nabla_\lambda \log q(\mbz ; \lambda)$.
We index the function class $\cF$ with parameters $\theta$, and
require that $f_\theta(\cdot)$ is differentiable. In the experiments,
we
use neural networks, which are flexible enough to approximate a
general family of test functions~\citep{Hornik:1989}.

Given parameterizations of the variational family and test family,
\gls{OPVI} seeks to solve a minimax problem,
\begin{align}
  \mblambda^* = \inf_{\mblambda} \, \sup_{\mbtheta} \,
  t(\E_{\mblambda} [(O^{p,q} f_{\mbtheta})(\mbz)]).
\end{align}
We will use stochastic optimization~\citep{Robbins:1951,Kushner:1997}.
In principle, we can find stochastic gradients of $\mblambda$ by
rewriting the objective in terms of the optimized value of $\mbtheta$,
$\mbtheta^*(\mblambda)$. In practice, however, we simultaneously solve
the maximization and minimization. Though computationally
beneficial, this produces saddle points. In our experiments we found
it to be stable enough.
We derive gradients
for the variational parameters $\mblambda$ and test function
parameters $\mbtheta$.  (We fix the distance function to be
the square $t(a) = a^2$; the identity $t(a)=a$ also readily
applies.)

\parhead{Gradient with respect to $\mblambda$.}  For a fixed test
function with parameters $\mbtheta$, denote the objective
$$\cL_{\mbtheta} = t(\E_{\mblambda} [(O^{p,q} \, f_{\mbtheta})(\mbz)]).$$  The
gradient with respect to variational parameters $\mblambda$ is
\begin{align*}
  \nabla_{\mblambda}  \cL_{\mbtheta} =
  2~\E_{\mblambda} [(O^{p,q} \, f_{\mbtheta})(\mbz)]~\nabla_{\mblambda}
  \E_{\mblambda} [(O^{p,q} \, f_{\mbtheta})(\mbz)].
\end{align*}
Now write the second expectation with the score function
gradient~\citep{Ranganath:2014}. This gradient is
\begin{align}
  \nabla_{\mblambda}  \cL_{\mbtheta} =
  2~\E_{\mblambda} [(O^{p,q} \, f_{\mbtheta})(\mbz)]~\E_{\mblambda}
  [\nabla_{\mblambda} \log q(\mbz ; \mblambda) (O^{p,q} \,
  f_{\mbtheta})(\mbz) + \nabla_{\mblambda} (O^{p,q} \, f_{\mbtheta})(\mbz)].
  \label{eq:grad-lambda}
\end{align}
\myeqp{grad-lambda} lets us calculate unbiased stochastic
gradients. We first generate two sets of independent samples from
$q$; we then form Monte Carlo estimates of the first and second
expectations. For the second expectation, we can use the variance
reduction techniques developed for black box variational inference,
such as Rao-Blackwellization~\citep{Ranganath:2014}.

We described the score gradient because it is general.  An alternative
is to use the reparameterization gradient for the second
expectation~\citep{Kingma:2014,Rezende:2014}.  It requires that the
operator be differentiable with respect to $\mbz$ and that samples
from $q$ can be drawn as a transformation $r$ of a parameter-free
noise source $\mbepsilon$, $\mbz = r(\mbepsilon, \mblambda)$.  In our
experiments, we use the reparameterization gradient.

\parhead{Gradient with respect to $\mbtheta$.}
Mirroring the notation above, the operator objective for fixed
variational $\mblambda$ is
\begin{align*}
  \cL_{\mblambda} = t(\E_{\mblambda} [(O^{p,q} \, f_{\mbtheta})(\mbz)]).
\end{align*}
The gradient with respect to test function parameters $\mbtheta$ is
\begin{align}
  \nabla_{\mbtheta}  \cL_{\mblambda} = 2~\E_{\mblambda} [(O^{p,q}
  f_{\mbtheta})(\mbz)]~\E_{\mblambda}[\nabla_{\mbtheta} O^{p,q} \, f_{\mbtheta}(\mbz)].
\label{eq:grad-theta}
\end{align}
Again, we can construct unbiased stochastic gradients with two sets of
Monte Carlo estimates.
Note that gradients for the test function do not require score
gradients (or reparameterization gradients) because the expectation
does not depend on $\mbtheta$.

\parhead{Algorithm.}  \myalg{operator_vi} outlines \gls{OPVI}.  We
simultaneously minimize the variational objective with respect to the
variational family $q_{\mblambda}$ while maximizing it with respect to the
function class $f_{\mbtheta}$.  Given a model, operator, and function
class parameterization, we can use automatic differentiation to
calculate the necessary gradients~\citep{carpenter2015stan}.  Provided
the operator does not require model-specific computation, this
algorithm satisfies the black box criteria.

\begin{algorithm}[t]
\SetKwInOut{Input}{Input}
\SetKwInOut{Output}{Output}
\Input{Model $\log p(\mbx, \mbz)$, variational approximation $q(z ; \mblambda)$}
\Output{Variational parameters $\mblambda$}
Initialize $\mblambda$ and $\mbtheta$ randomly. \\
\While{not converged}{
  Compute unbiased estimates of $\nabla_{\mblambda} \cL_{\mbtheta}$ from
  \myeqp{grad-lambda}.\\
  Compute unbiased esimates of $\nabla_{\mbtheta} \cL_{\mblambda}$ from
  \myeqp{grad-theta}. \\
  Update $\mblambda$, $\mbtheta$ with unbiased stochastic gradients.\\
 }
 \caption{\Glsfirst{OPVI}}
 \label{alg:operator_vi}
\end{algorithm}

\subsection{Data Subsampling and \gls{OPVI}}

With stochastic optimization, data subsampling scales up
traditional variational inference to massive
data~\citep{Hoffman:2013,Titsias:2014}.  The idea is to calculate
noisy gradients by repeatedly subsampling from the data set,
without needing to pass through the entire data set for each
gradient.

An as illustration, consider hierarchical models. Hierarchical models consist of global latent
variables $\beta$ that are shared across data points and local latent
variables $z_i$ each of which is associated to a data point $x_i$.
The model's log joint density is
\begin{align*}
  \log p(x_{1:n}, z_{1:n}, \beta)
  = \log p(\beta) +
  \sum_{i=1}^n\Big[\log p(x_i \g z_i, \beta) + \log p(z_i \g \beta)\Big].
\end{align*}
\citet{Hoffman:2013} calculate unbiased estimates of the log joint
density (and its gradient) by subsampling data and appropriately
scaling the sum.

We can characterize whether \gls{OPVI} with a particular operator
supports data subsampling.  \gls{OPVI} relies on evaluating the
operator and its gradient at different realizations of the latent
variables (\myeqp{grad-lambda} and \myeqp{grad-theta}). Thus we
can subsample data to calculate estimates of the operator when it
derives from linear operators of the log density, such as
differentiation and the identity.  This follows as a linear operator
of sums is a sum of linear operators, so the gradients in
~\myeqp{grad-lambda} and~\myeqp{grad-theta} decompose into a sum. The \acrlong{LS}
and \acrshort{KL} operator
are both linear in the log density; both
support data subsampling.

\subsection{Variational Programs}

Given an operator and variational family, \myalg{operator_vi} optimizes the
corresponding operator objective.
Certain operators require the density of $q$.  For example, the
\acrshort{KL} operator (\myeqp{KL-operator}) requires its log
density.  This potentially limits the construction of rich variational
approximations for which the density of $q$ is difficult to
compute.\footnote{It is possible to construct rich approximating
  families with $\textsc{kl}(q || p)$, but this requires the introduction
  of an auxiliary distribution~\citep{maaloe2016auxiliary}.}

Some
operators, however, do not depend on having a analytic density; the
\gls{LS} operator (\myeqp{lang-stein}) is an example.  These
operators can be used with a much richer class of variational
approximations, those that can be sampled from but might not have
analytically tractable densities.  We call such approximating families
\emph{variational programs}.

Inference with a variational program requires the family to be
reparameterizable~\citep{Kingma:2014,Rezende:2014}.  (Otherwise we
need to use the score function, which requires the derivative of the
density.) A reparameterizable variational program consists of a
parametric deterministic transformation $R$ of random noise $\mbepsilon$. Formally, let
\begin{align}
\mbepsilon \sim \textrm{Normal}(0, 1), \quad \mbz = R(\mbepsilon; \mblambda).
\label{eq:reparam-var-program}
\end{align}
This generates samples for $\mbz$, is differentiable with respect to
$\mblambda$, and its density may be intractable. For operators that do
not require the density of $q$, it can be used as a powerful variational
approximation. This is in contrast to the standard \gls{KL} operator.

As an example, consider the following variational program
for a one-dimensional random variable. Let $\lambda_i$ denote the $i$th
dimension of $\mblambda$ and make the corresponding definition for $\mbepsilon$:
\begin{align}
z = (\epsilon_3 > 0) R(\epsilon_1; \lambda_1) - (\epsilon_3 \leq 0) R(\epsilon_2; \lambda_2).
\label{eq:toy_var_prog}
\end{align}
When $R$ outputs positive values, this separates the parametrization
of the density to the positive and negative halves of the reals; its
density is generally intractable.  In \mysec{experiments}, we will
use this distribution as a variational approximation.

\myeqp{reparam-var-program} contains many densities when the function
class $R$ can approximate arbitrary continuous functions.  We state it
formally.  \vspace{0.05in}
\begin{theorem}
  Consider a posterior distribution $p(\mbz\g\mbx)$ with a finite
  number of latent variables and continuous quantile function.  Assume
  the operator variational objective has a unique root at the
  posterior $p(\mbz\g\mbx)$ and that $R$ can approximate continuous
  functions.  Then there exists a sequence of parameters
  $\lambda_1,\lambda_2\ldots,$ in the variational program, such that
  the operator variational objective converges to $0$, and thus $q$
  converges in distribution to $p(\mbz\g\mbx)$.
\end{theorem}
This theorem says that we can use variational programs with an
appropriate $q$-independent operator to approximate continuous
distributions.  The proof is in Appendix D.

\section{Empirical Study}
\label{sec:experiments}
\glsresetall

We evaluate operator variational inference on a mixture of Gaussians,
comparing different choices in the objective. We then study logistic
factor analysis for images.

\subsection{Mixture of Gaussians}
Consider a one-dimensional
mixture of Gaussians as the posterior of interest,
$p(z)~=~\frac{1}{2} \textrm{Normal}(z; -3, 1) + \frac{1}{2}
\textrm{Normal}(z; 3, 1)$.
The posterior contains multiple modes.  We seek to approximate it
with three variational objectives: \gls{KL} with a Gaussian
approximating family, \gls{LS} with a Gaussian approximating family, and \gls{LS}
with a variational program.

\begin{figure}[ht]
\centering
\begin{subfigure}{0.23\linewidth}
  \centering
  \includegraphics[trim=0 0 0 0, width=\linewidth]{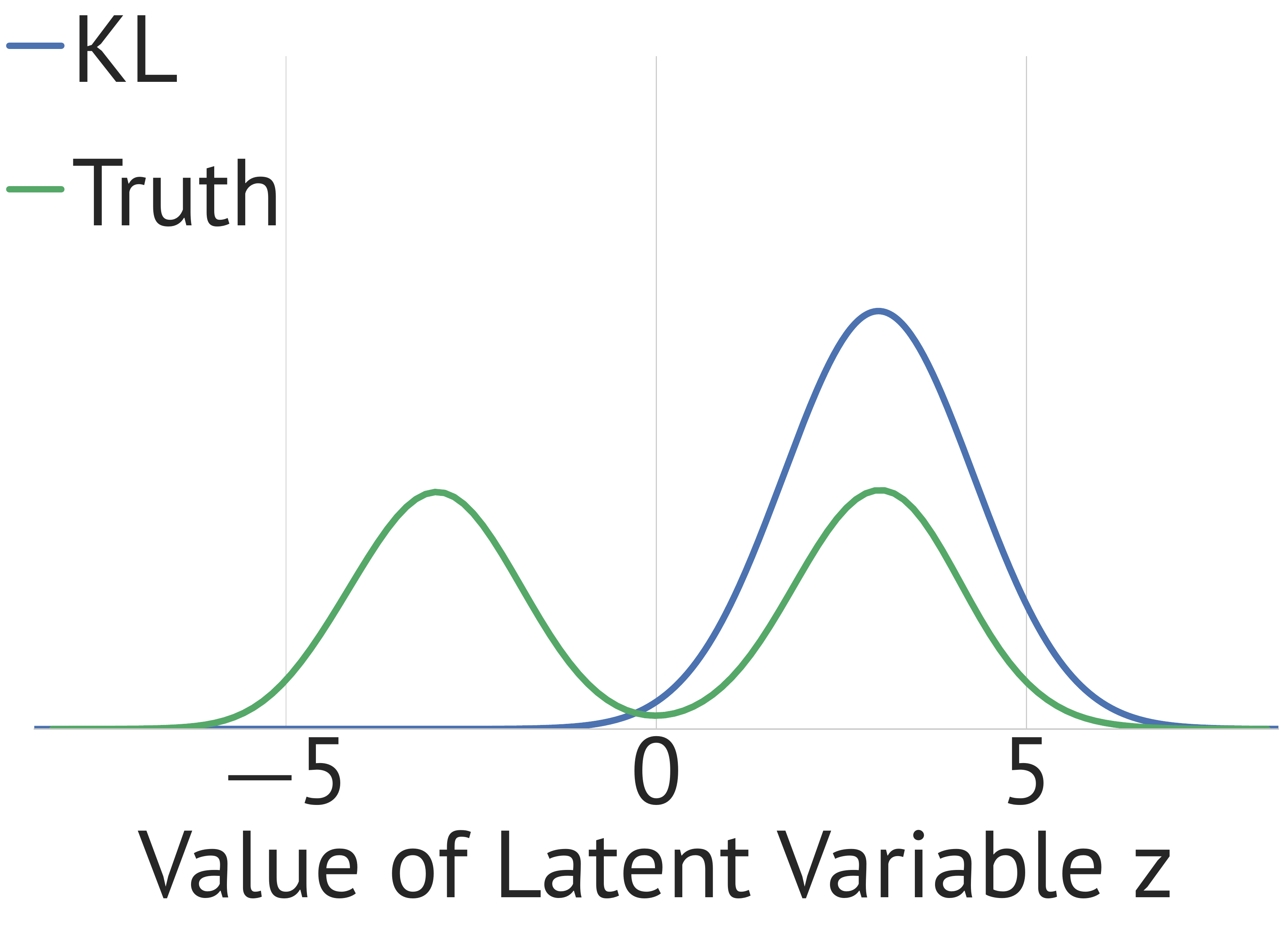}
\end{subfigure}
\begin{subfigure}{0.23\textwidth}
  \centering
  \includegraphics[trim=0 0 0 0, width=\linewidth]{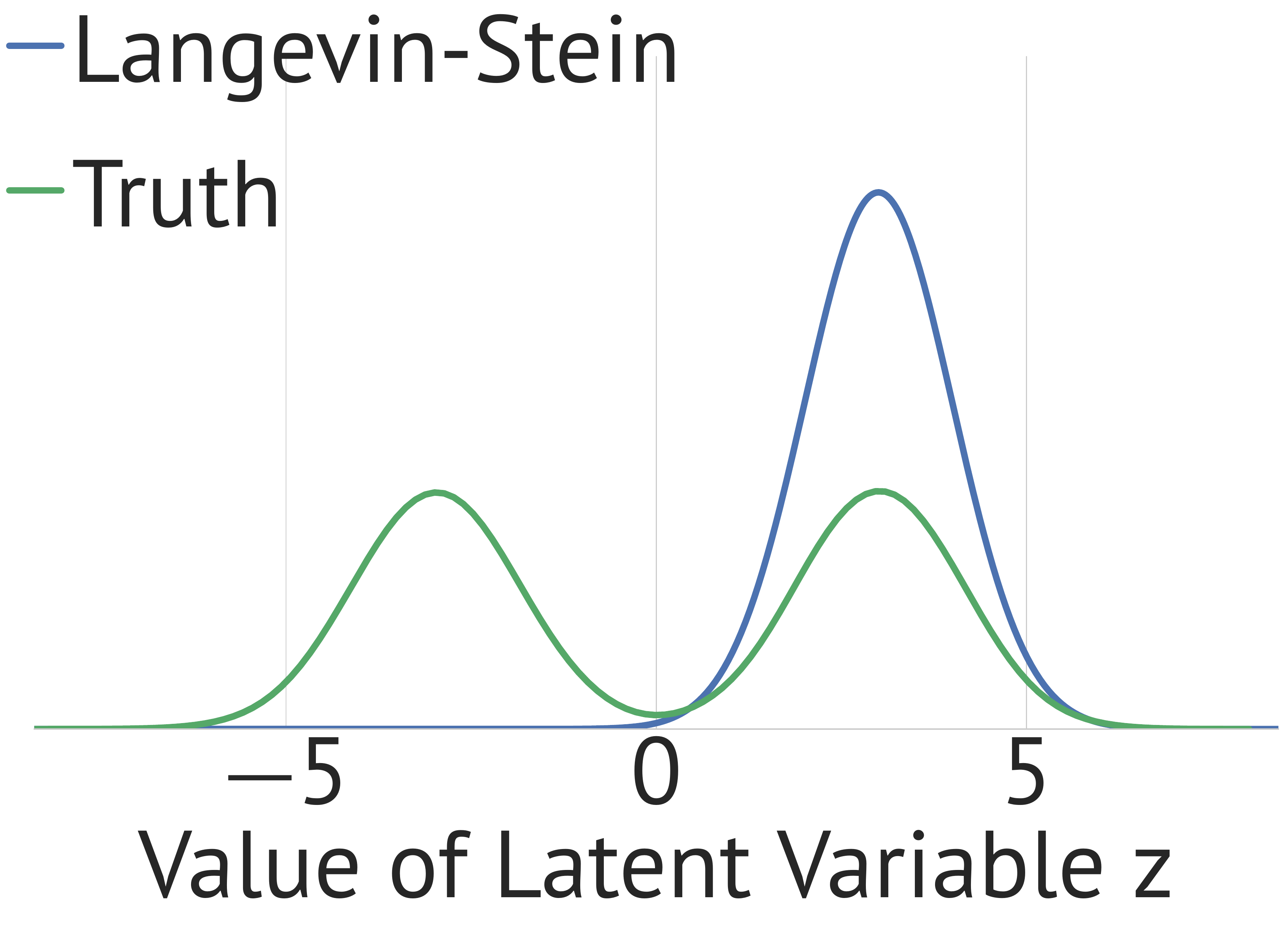}
\end{subfigure}
\begin{subfigure}{0.23\textwidth}
    \centering
   \includegraphics[trim=0 0 0 0, width=\linewidth]{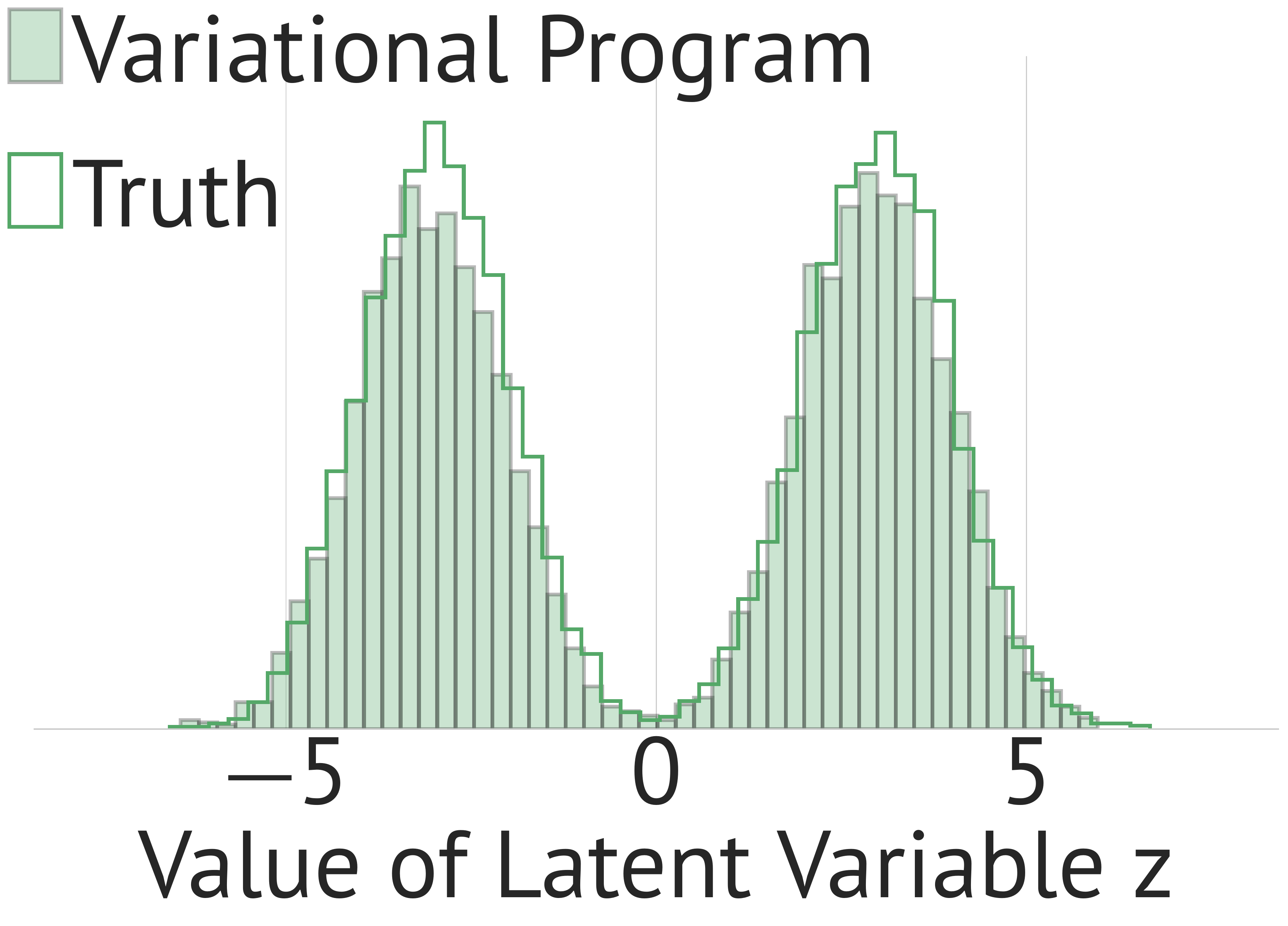}
\end{subfigure}
\caption{The true posterior is a mixture of two Gaussians, in green. We approximate it with
  a Gaussian using two operators (in blue). The density on the far
  right is a variational program given in~\myeqp{toy_var_prog} and
  using the Langevin-Stein operator; it approximates the truth well.
  The density of the variational program is intractable. We plot a histogram
  of its samples and compare this to the histogram of the true posterior.}
\label{fig:mog}\label{fig:toy}
\end{figure}

\myfig{mog} displays the posterior approximations.  We find that the \gls{KL}
divergence and \gls{LS} divergence choose a single mode and have
slightly different variances. These operators do not produce good
results because a single Gaussian is a poor approximation to the
mixture. The remaining distribution in~\myfig{mog} comes from the
toy variational program described by~\myeqp{toy_var_prog} with the
\gls{LS} operator.  Because this program
captures different distributions for the positive and negative half of the
real line, it is able to capture the posterior.

In general, the choice of an objective balances statistical and computational properties of variational inference. We highlight one tradeoff: the \gls{LS} objective admits the use of a variational program; however, the objective is more difficult to optimize than the \gls{KL}.

\subsection{Logistic Factor Analysis}
Logistic factor analysis models binary vectors $\mbx_i$ with a matrix of
parameters $\mbW$ and biases $\mbb$,
\begin{align*}
  \mbz_i &\sim \textrm{Normal}(0, 1) \\
  x_{i,k} &\sim \textrm{Bernoulli}(\sigma(\mbw_k^\top \mbz_i + b_k)),
\end{align*}
where $\mbz_i$ has fixed dimension $K$ and $\sigma$ is the sigmoid
function.  This model captures correlations of the entries in $\mbx_i$
through $\mbW$.

We apply logistic factor analysis to analyze the binarized MNIST data
set~\citep{salakhutdinov2008quantitative},
which contains 28x28 binary pixel images of handwritten digits.
(We set the latent dimensionality to 10.)  We fix the model parameters
to those learned with variational expectation-maximization using the
\gls{KL} divergence, and focus on comparing posterior inferences.

We compare the \gls{KL} operator to the \gls{LS} operator and study
two choices of variational models: a fully factorized Gaussian
distribution and a variational program. The variational program
generates samples by transforming a $K$-dimensional standard normal
input with a two-layer neural network, using rectified linear activation
functions and a hidden size of twice the latent dimensionality. Formally, the variational
program we use generates samples of $\mbz$ as follows:
\begin{align*}
\mbz_0 &\sim \textrm{Normal}(0, I) \\
\mbh_0 &= \textrm{ReLU}({\mbW_0^q}^\top \mbz_0 + \mbb_0^q) \\
\mbh_1 &= \textrm{ReLU}({\mbW_1^q}^\top \mbh_0 + \mbb_1^q) \\
\mbz &= {\mbW_2^q}^\top \mbh_1 + \mbb_2^q.
\end{align*}
The variational parameters are the weights $\mbW^q$ and biases $\mbb^q$.
For $f$, we use a
three-layer neural network with the same hidden size as the
variational program and hyperbolic tangent activations where unit
activations were bounded to have norm two. Bounding the unit norm
bounds the divergence. We used the Adam optimizer \citep{kingma2014adam} with learning rates $2 \times 10^{-4}$ for
$f$ and $2 \times 10^{-5}$ for the variational approximation.

\begin{table}[tb]
\centering
\begin{tabular}{lcc}
\toprule
Inference method & Completed data log-likelihood
\\
\midrule
Mean-field Gaussian + \gls{KL} & -59.3 \\
Mean-field Gaussian + \gls{LS} &  -75.3 \\
Variational Program + \gls{LS} & -58.9\\
\bottomrule
\end{tabular}
\vspace{1ex}
\caption{Benchmarks on logistic factor analysis for binarized MNIST.
The same variational approximation with \gls{LS} performs worse than
\gls{KL} on likelihood performance. The variational program with
\gls{LS} performs better without directly optimizing for likelihoods.}
\label{table:mnist}
\vskip -.1in
\end{table}

There is no standard for evaluating generative models and their
inference algorithms~\citep{theis2016note}.
Following
\citet{Rezende:2014}, we consider a missing data problem. We remove
half of the pixels in the test set (at random) and reconstruct them
from a fitted posterior predictive distribution. \mytable{mnist}
summarizes the results on 100 test images; we report the
log-likelihood of the completed image. \gls{LS} with the variational
program performs best.  It is followed by \gls{KL} and the simpler
\gls{LS} inference. The \gls{LS} performs better than \gls{KL} even
though the model parameters were learned with \gls{KL}.

\section{Summary}
\label{sec:discussion}

We present operator variational objectives, a broad yet tractable
class of optimization problems for approximating posterior
distributions.  Operator objectives are built from an operator, a
family of test functions, and a distance function.  We outline the
connection between operator objectives and existing divergences such
as the KL divergence, and develop a new variational objective using
the Langevin-Stein operator.  In general, operator objectives produce
new ways of posing variational inference.

Given an operator objective, we develop a black box algorithm for
optimizing it and show which operators allow scalable optimization
through data subsampling.  Further, unlike the popular evidence lower
bound, not all operators explicitly depend on the approximating
density. This permits flexible approximating families, called
variational programs, where the distributional form is not
tractable. We demonstrate this approach on a mixture model and a
factor model of images.

There are several possible avenues for future directions such as
developing new variational objectives, adversarially learning~\citep{goodfellow2014generative} model parameters with operators, and
learning model parameters with operator variational objectives.

\parhead{Acknowledgments.}
This work is supported by NSF IIS-1247664,  ONR
N00014-11-1-0651, DARPA FA8750-14-2-0009, DARPA
N66001-15-C-4032, Adobe, NSERC PGS-D, Porter Ogden Jacobus Fellowship,
Seibel Foundation, and the Sloan Foundation. The authors would
like to thank Dawen Liang, Ben Poole, Stephan Mandt, Kevin Murphy, Christian
Naesseth,
and the anonymous reviews for their helpful feedback and comments.

\section*{References}
\renewcommand{\bibsection}{}
\bibliographystyle{apalike}
{\small
\bibliography{bib}
}

\appendix

\section{Technical Conditions for Langevin-Stein Operators}
\label{sec:zero_conditions}
Here we establish the conditions needed on the function class $\cF$ or the
posterior distribution shorthanded $p$ for the operators to have expectation zero
for all $f \in \cF$. W derive properties using integration by parts
for supports that are bounded open sets. Then we extend the result
to unbounded supports using limits. We start with the Langevin-Stein operator. Let $S$ be the
set over which we integrate and let $B$ be its boundary. Let $v$ be the unit normal to
the surface $B$, and $v_i$ be the $i$th component of the surface normal (which is
$d$ dimensional). Then we have that
\begin{align*}
\int_S p  &(O^{p}_\textsc{LS}  \, f) dS =  \int_S p \nabla_z \log p^\top f + p \nabla^\top f dS
\\
&= \sum_{i=1}^d \int_S \frac{\partial}{\partial{z_i}}[p] f_i + p \frac{\partial}{\partial{z_i}}[f_i]dS
\\
&= \sum_{i=1}^d \int_S \frac{\partial}{\partial{z_i}}[p] f_idS + \int_B f_i p v_i dB - \int_S \frac{\partial}{\partial{z_i}}[p] f_idS
\\
&= \int_B v^\top f p dB.
\end{align*}
A sufficient condition for this expectation to be zero is that either $p$ goes to zero at its boundary or that the vector field $f$ is zero at the boundary.

For unbounded sets, the result can be written as a limit for a sequence of increasing sets $S_n \to S$ and a set of boundaries $B_n \to B$ using the dominated convergence theorem~\citep{Cinlar:2011}.
To use dominated convergence, we establish absolute integrability. Sufficient conditions for absolute integrability of the Langevin-Stein operator are for the gradient of $\log p$ to be bounded and the vector field $f$ and its derivatives to be bounded. Via dominated convergence, we get that $\lim_n \int_{B_n} v^\top f p dB = 0$ for the Langevin-Stein operator to have expectation zero.

\section{Characterizing the zeros of the Langevin-Stein Operators}
\label{sec:optimal_operator}
We provide analysis on how to characterize the equivalence class of
distributions defined as $(O^{p,q}f)(z) = 0$. One general condition for
equality in distribution comes from equality in probability on all Borel sets.
We can build functions that have expectation zero with respect to the posterior
that test this equality.
Formally, for any Borel set $A$ with $\delta_A$ being the indicator, these functions on $A$ have the form:
\begin{align*}
\delta_{A}(\mbz) - \int_A p(\mby) d\mby
\end{align*}
We show that if the Langevin-Stein operator
satisfies $\cL(q ; O^{p}_\textsc{LS}, \cF) = 0$, then $q$ is equivalent to $p$
in distribution. We do this by showing the above functions are in the span of $O^{p}_\textsc{LS}$.
Expanding the Langevin-Stein operator we have
\begin{align*}
(O^{p}_\textsc{LS}  \, f)  =  p^{-1} \nabla_z p^\top f + \nabla^\top f
= p^{-1} \sum_{i=1}^d  \frac{\partial{f_i p}}{{\partial z_i}}.
\end{align*}
Setting this equal to the desired function above yields the differential equation
\begin{align*}
\delta_{A}(z) -  \int_A p(y) dy
= p^{-1}(z) \sum_{i=1}^d  \frac{\partial{f_i p}}{{\partial z_i}}(z).
\end{align*}
To solve this, set $f_i = 0$ for all but $i=1$. This yields
\begin{align*}
\delta_{A}(z) -  \int_A p(y) dy
= p^{-1}(z) \frac{\partial{f_1 p}}{{\partial z_1}}(z),
\end{align*}
which is an ordinary differential equation with solution for $f_1$
\begin{align*}
f_1^A(z) = \frac{1}{p(z)} \int\limits_{-\infty}^{z_1} p(a, z_{2...d}) \left(\delta_{A}(a, z_{2...d}) -  \int_A p(\mby) d\mby \right) da.
\end{align*}
This function is differentiable with respect to $z_1$, so this gives the desired result. Plugging the function back into the operator variational objective gives
\begin{align*}
\E_q\left[\delta_{A}(\mbz) -  \int_A p(\mby) d\mby\right] = 0 \iff \E_q[\delta_{A}(\mbz)] = \E_p[\delta_{A}(\mbz)],
\end{align*}
for all Borel measurable $A$. This implies the induced distance captures total variation.

\section{Operators for Discrete Variables}
\label{sec:discrete_vars}
Some operators based on Stein's method
are applicable only for latent variables in a continuous space. There are Stein
operators that work with discrete variables~\citep{Assaraf:1999,Ley:2011b}.
We present one amenable to operator variational objectives based on a discrete analogue to the Langevin-Stein operator developed in~\cite{Ley:2011b}. For simplicity, consider a one-dimensional
discrete posterior with support $\{0, ..., c\}$. Let $f$ be a function
such that $f(0) = 0$, then an operator can be defined as
\begin{align*}
(O^{p}_\textsc{discrete} \, f)(z) = \frac{f(z + 1) p(z + 1, \mbx) - f(z) p(z , \mbx)}{p(z, \mbx)}.
\end{align*}
Since the expectation of this operator with respect to the posterior $p(z \g x)$
is a telescoping sum with both endpoints $0$,
it has expectation zero.

This relates to the Langevin-Stein operator in the following. The Langevin-Stein operator in one dimension can be written as
\begin{align*}
(O^{p}_\textsc{LS}  \, f)  =  \frac{\frac{d}{dz}[f p]}{p}.
\end{align*}
This operator is the discrete analogue as the differential is replaced by
a discrete difference. We can extend this operator to multiple dimensions
by an ordered indexing. For example, binary numbers of length $n$ would work for $n$ binary
latent variables.

\section{Proof of Universal Representations}
\label{sec:universal_representation}

Consider the optimal form of $R$ such that transformations of standard
normal draws are equal in distribution to exact draws from the
posterior. This means
\begin{equation*}
R(\mbepsilon;\mblambda) = P^{-1}(\Phi(\mbepsilon)),
\end{equation*}
where $\Phi(\mbepsilon)$ squashes the draw from a standard normal such that it
is equal in distribution to a uniform random variable. The posterior's inverse cumulative distribution function $P^{-1}$ is applied to the uniform draws. The transformed samples are now equivalent to exact samples
from the posterior. For a rich-enough parameterization of $R$, we may hope
to sufficiently approximate this function.

Indeed, as in
the universal approximation theorem of \citet{tran2016variational}
there exists a sequence of parameters $\{\lambda_1,\lambda_2,\ldots\}$ such that the operator
variational objective goes to zero, but the function class is no longer limited
to local interpolation. Universal approximators like neural networks~\citep{Hornik:1989}
also work. Further, under the assumption that $p$ is the unique root and by
satisfying the conditions described in \mysec{optimal_operator} for equality in
distribution, this implies that the variational program given by
drawing $\mbepsilon\sim\mathcal{N}(\mb{0},\mbI)$ and applying
$R(\mbepsilon)$ converges in distribution to $p(\mbz\g\mbx)$.

\end{document}